%% file: main.tex
\documentclass[letterpaper, 10 pt, conference]{ieeeconf}  

\IEEEoverridecommandlockouts                              

\usepackage{graphicx}
\usepackage{graphics}
\usepackage{color,xcolor}
\usepackage{amsmath,amsfonts}

\graphicspath{{figures/}}
\newcommand{\mbf}[1]{\mathbf{#1}}
\newcommand{\bR}{\mathbb{R}}

\DeclareMathOperator{\reach}{reach}
\DeclareMathOperator{\wfs}{wfs}

\title{\LARGE \bf
Output Mode Switching for Parallel Five-bar Manipulators \\
Using a Graph-based Path Planner
}

\author{Parker B. Edwards$^{1}$, Aravind Baskar$^{2}$, Caroline Hills$^{1}$, Mark Plecnik$^{2}$*, 
and Jonathan D. Hauenstein$^{1}$
\thanks{This material was supported in part by the National Science Foundation under Grant Nos. CMMI-2041789, CMMI-2144732, and CCF-1812746.}
\thanks{$^{1}$Parker B. Edwards, Caroline Hills, and Jonathan D. Hauenstein 
    are with Department of Applied and Computational Mathematics and Statistics, University of Notre Dame, Notre Dame, IN 46556 USA
        {\tt\small \{parker.edwards,chills1,hauenstein\}@nd.edu}}%
\thanks{$^{2}$Aravind Baskar and Mark Plecnik are with Department of Aerospace and Mechanical Engineering,
        University of Notre Dame, Notre Dame, IN 46556 USA
        {\tt\small \{abaskar,plecnikmark\}@nd.edu}}%
}

\begin{document}

\maketitle
\thispagestyle{empty}
\pagestyle{empty}

\begin{abstract}

The configuration manifolds of parallel manipulators exhibit more nonlinearity than serial manipulators.
Qualitatively, they can be seen to possess extra folds.
By projecting such manifolds onto spaces of engineering relevance, such as an output workspace or an input actuator space, these folds cast edges that exhibit nonsmooth behavior.
For example, inside the global workspace bounds of a five-bar linkage appear several local workspace bounds that only constrain certain output modes of the mechanism.
The presence of such boundaries, which manifest in both input and output projections, serve as a source of confusion when these projections are studied exclusively instead of the configuration manifold itself.
Particularly, the design of nonsymmetric parallel manipulators has been confounded by the presence of exotic projections in their input and output spaces.
In this paper, we represent the configuration space with a radius graph, then weight each edge by solving an optimization problem using homotopy continuation to quantify transmission quality.
We then employ a graph path planner to approximate geodesics between configuration points that avoid regions of low transmission quality.
Our methodology automatically generates paths capable of transitioning between non-neighboring output modes, a motion which involves osculating multiple workspace boundaries (local, global, or both).
We apply our technique to two nonsymmetric five-bar examples that demonstrate how transmission properties and other characteristics of the workspace can be selected by switching output modes.

\end{abstract}

\section{INTRODUCTION}\label{Sec:Intro}

\begin{figure}[t]
	\centering
	\includegraphics[width=0.48\textwidth]{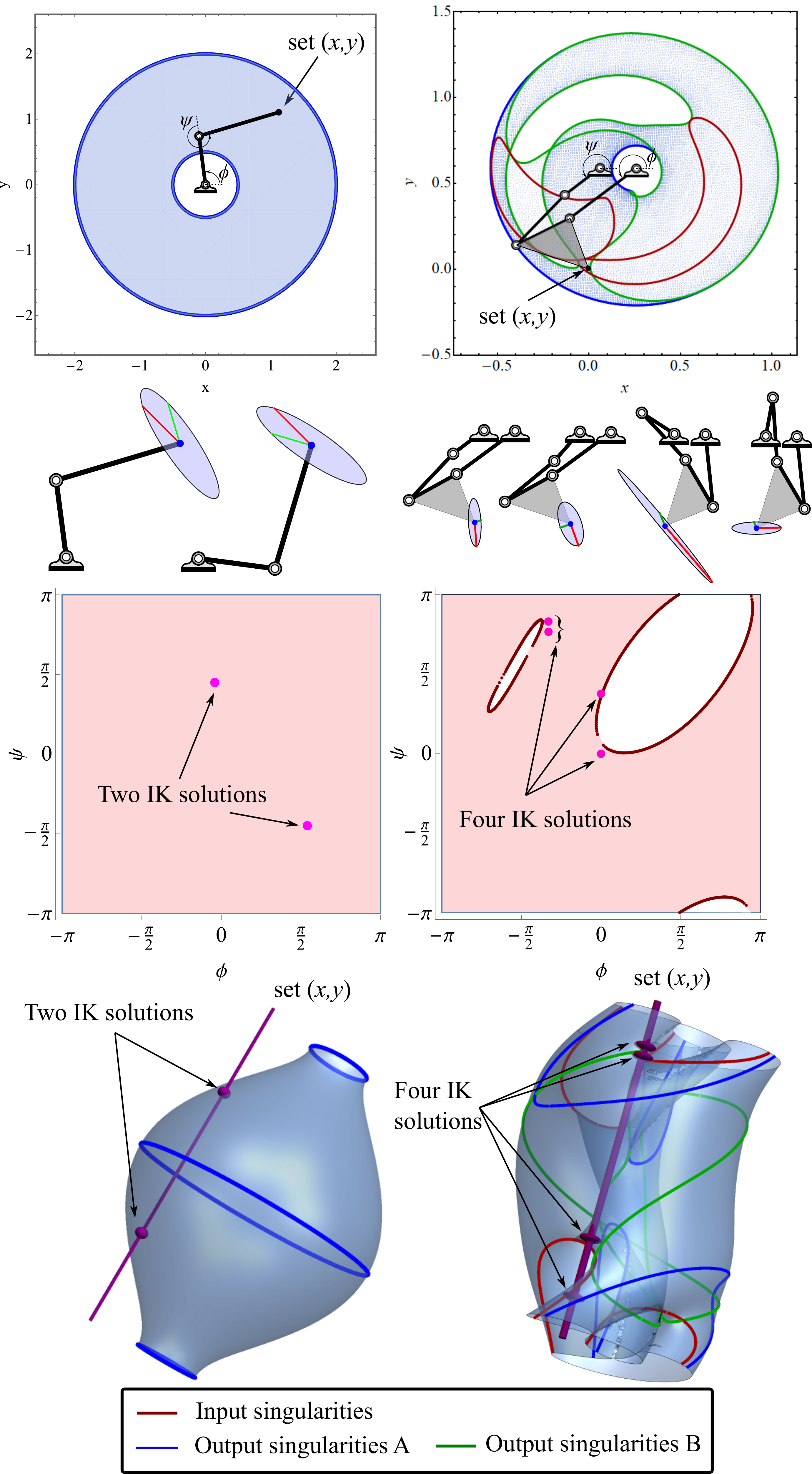}
	\vspace{-6mm}
	\caption{
	Views of the configuration spaces of a serial 2R (left half) and a parallel five-bar (right half).
	1st row: The output workspace for each manipulator.
	2nd row: The 2R admits two IK solutions, the five-bar admits four.
	3rd row: The input actuator space for each manipulator, with holes visible for the five-bar.
	4th row: An $(x, y, \phi)$ view of the configuration manifold that shows smoothness everywhere.
	The skewering line shows the locations of the IK solutions for a set $(x, y)$ .
	}
	\label{fig:schematic_intro}
	\vspace{-4mm}
\end{figure}

The workspaces of parallel mechanisms are more complicated than their serial counterparts.
Their interiors generically contain inner bounds of different sorts: either favorable or unfavorable transmission characteristics, or locations where it is possible to transition between different modes of the mechanism.
Dealing with such complexity is a burden, but provides opportunity for kinematical advantages.
For example, at a single output point, parallel mechanisms can assume a larger selection of configurations that afford more choice of directional transmission characteristics.
To benefit from such choice, there needs 
to be a plan for transitioning between these different modes, which is the contribution of this work.
To explain the complexity we plan to navigate, we compare two planar two degree-of-freedom linkages: the serial 2R (Fig.~\ref{fig:schematic_intro} left) and the parallel five-bar (Fig.~\ref{fig:schematic_intro} right).

The workspace of the 2R is an annulus.
The linkage can assume two different configurations for every point within its workspace. 
To transition between these different modes, the end-effector point must travel to a workspace bound and come back.
Placing a rotary actuator at each joint, there are no points in the workspace where the linkage is incapable of exerting a force in any direction.
At each point, the selection of configurations and velocity ellipses \cite{lynchModernRoboticsMechanics2017} available are reflections of one another, see Fig.~\ref{fig:schematic_intro}.

The workspace bounds of a five-bar is a union of two circular arc segments and two four-bar coupler curves.
Despite having two degrees-of-freedom,
its configuration space cannot be fully visualized with only two parameters.
Therefore, viewing the $x$-$y$ projection of its output point (i.e., its Cartesian workspace) seemingly shows additional interior workspace bounds.
To add to the confusion, each internal curve of Fig.~\ref{fig:schematic_intro} is relevant only to certain output modes and irrelevant to others.
Some of these bounds indicate holes analogous to the annulus
while others indicate bounds of specific output modes 
which serve as places of transition between modes.
Scattered within its configuration space are curves at which this linkage loses its ability to exert forces in a direction.
These curves as well are only relevant to some output modes but not others.
For a regular workspace point, the linkage can assume two or four different configurations, with each exhibiting a different velocity ellipse.
To study these configuration spaces, we use the terms \emph{input singularities}, \emph{output singularities}, \emph{input modes}, and \emph{output modes}.

\begin{figure}[b]
	\centering
	\includegraphics[width=0.3\textwidth]{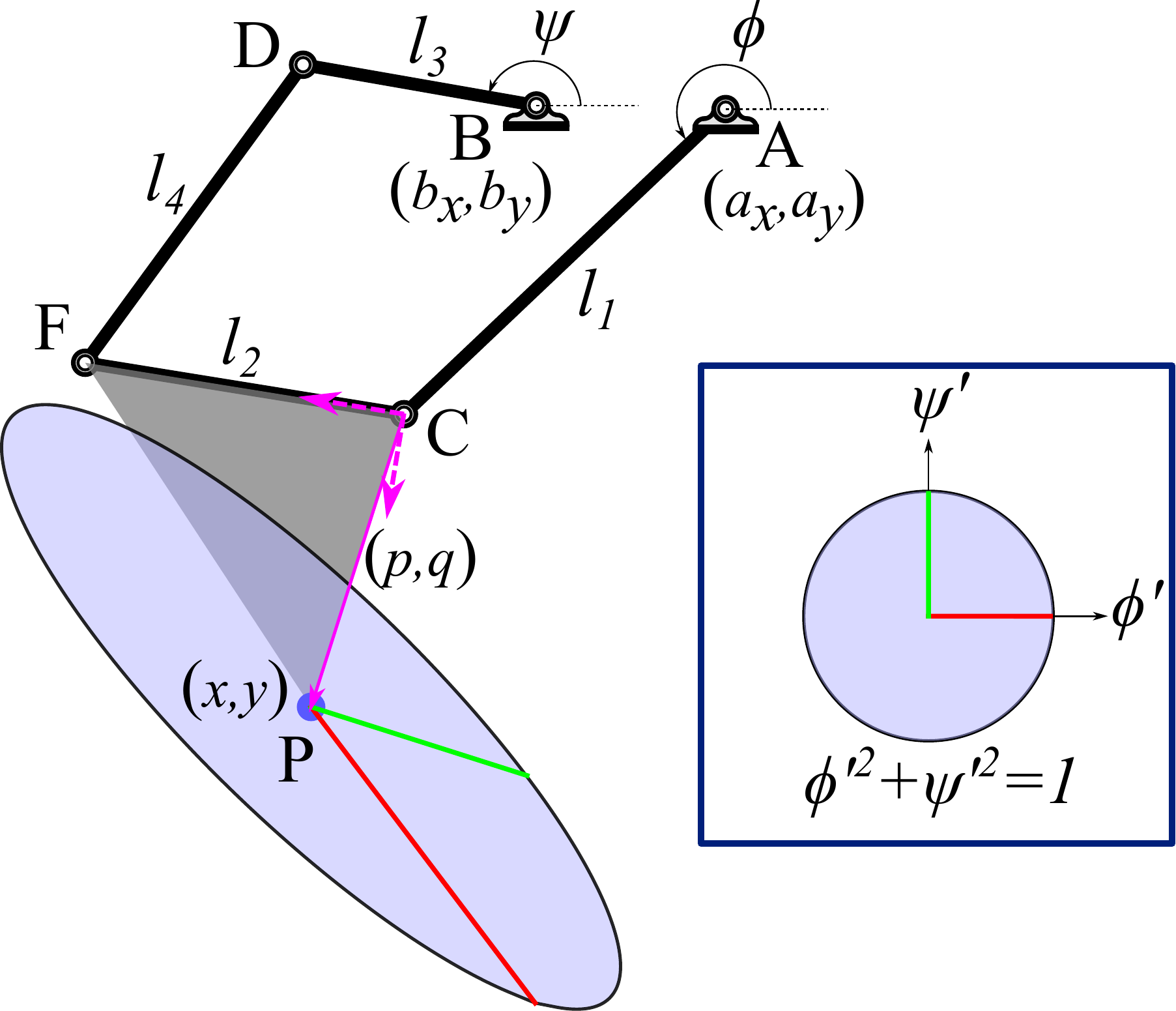}
	\caption{Schematic of a five-bar mechanism. The velocity ellipse with respect to the input angles~$\phi$ and~$\psi$ is shown at its end-effector~$(x,y)$.}
	\label{fig:schematic}
\end{figure}

\textbf{Input singularities}.
Input singularities are the singular solutions to the forward kinematics problem.
In the $\phi$-$\psi$ plane (input space, Fig.~\ref{fig:schematic_intro}), input singularities serve as bounds where \emph{locally} the mechanism can configure on one side but not the other.
At these bounds, one of the semi-axis lengths of the velocity ellipse in output space tends toward infinity.
It is in that direction that the linkage cannot transmit any force at the end-effector no matter the torque exerted by the actuators.
Because of this loss of control authority, input singularities are generally considered to have \textbf{transmission problems} and should be avoided.
For a five-bar linkage, the geometric condition for an input singularity is that points $C$, $D$, $F$, are collinear (Fig.~\ref{fig:schematic}).
Gosselin and Angeles \cite{gosselinSingularityAnalysisClosedloop1990} refer to these as Type II singularities.
Urizar et al.~\cite{urizarComputingConfigurationSpace2009} refer to these as DKP (direct kinematic problem) singularities.

\textbf{Output Singularities}.
Output singularities are the singular solutions to the inverse kinematics problem.
In the $x$-$y$ plane (output space), output singularities serve as bounds where \emph{locally} the mechanism can configure on one side but not the other.
At these bounds, the semi-axis of the velocity ellipse normal to these curves collapses to zero.
It is in that direction that the linkage can sustain any force at the end-effector with zero torque from the actuators.
In this way, there are \textbf{no transmission problems} at output singularities from the vantage of torque exerted by the actuators.
It is at these configurations that the linkage is able to transition output modes.
For a five-bar linkage, the geometric condition for an output singularity is that either points $A$, $C$, $P$, or points $B$, $D$, $F$, are collinear (Fig.~\ref{fig:schematic}).
%
%
Gosselin and Angeles~\cite{gosselinSingularityAnalysisClosedloop1990} refer to these as Type I singularities.
Urizar et al.~\cite{urizarComputingConfigurationSpace2009} refer to these as IKP (inverse kinematic problem) singularities.


\textbf{Input modes}.
The separated regions that result after partitioning the configuration space by its input singularities are called its input modes.
Note that if input singularities do not exist, there may be one or many input modes, depending on the connectivity of the unpartitioned configuration space.
An input mode is a maximal continuous region of configurations whereby a path can be formed between any two regular interior members without passing through an input singularity.
%
%
The term \emph{assembly mode} appears frequently in past literature.
An \emph{input mode} is not an \emph{assembly mode}.
As defined in the past, the total number of assembly modes is equal to the number of solutions to the forward kinematics problem, implying the solutions uniquely identify all assembly modes.
The total number of input modes could be less than, equal, or greater than the number of assembly modes.
One way that it may be less is when a path exists between two forward kinematics solutions that does not pass through an input singularity, hence they are part of the same input mode.
Such cases are well documented and illustrate the shortcomings of the term \emph{assembly mode}~\cite{hernandezDefiningConditionsNonsingular2009}.
The number of input modes is greater than the number of assembly modes when the unpartitioned configuration space already consists of disconnected regions which are further divided by the presence of input singularities.
Input modes are of great practical importance because they are continuous regions in which the actuators maintain full control authority.
Additionally, in practice, input singularities should be avoided by a margin because transmission characteristics deteriorate near them as well.

\textbf{Output modes}.
The separated regions that result after partitioning the configuration space by its output singularities are called its output modes.
Note that if output singularities do not exist, there may be one or many output modes, depending on the connectivity of the unpartitioned configuration space.
An output mode is a maximal continuous region of configurations whereby a path can be formed between any two regular interior members without passing through an output singularity.
The term \emph{working mode} appears frequently in past literature.
An \emph{output mode} is not a \emph{working mode}.
As defined in the past, the total number of working modes is equal to the number of solutions to the inverse kinematics problem, implying the solutions uniquely identify all working modes.
The total number of output modes could be less than, equal, or greater than the number of working modes.
One way that it may be less is when a path exists between two inverse kinematics solutions that does not pass through an output singularity, hence they are part of the same output mode.
The number of output modes is greater than the number of working modes when the unpartitioned configuration space already consists of disconnected regions which are further divided by the presence of output singularities.
%

\section{LITERATURE REVIEW}\label{Sec:Review}

The serial 2R is the default choice when designing an $(x,y)$-planar manipulator
since it has a larger workspace than any comparable five-bar and never suffers from input space boundaries.
When it comes to selecting a five-bar, either symmetric or parallelogram designs
\cite{camposDevelopmentFiveBarParallel2011,kenneallyDesignPrinciplesFamily2016,hubickiATRIASDesignValidation2016,tianFlexurebasedFivebarMechanism2009}
seem to be popular with both having almost the same workspace and 
transmission characteristics as the serial 2R.
Grossly breaking this symmetry releases the complexity explained above.
We surmise that within the added complexity lies the opportunity for torque and energy savings.
For example, an internal output bound (which exhibits favorable torque transmission characteristics in its normal direction) can be shaped via dimensional synthesis to match a repeating trajectory (e.g., a walking gait) in order to reduce torque required from the actuators.
Another potential application is designing a manipulator that possesses an input mode that gives access to many IKP solutions which exhibit transmission characteristics advantageously biased toward either force production or backdrivability.
Appropriate output modes could be transitioned at will.  However, such a design effort is out of scope of this work.
Instead, this paper focuses on the prerequisite work to ensure such a manipulator can transition between different modes and 
plan~\mbox{its~path~accordingly}.
Macho et al.~\cite{machoWorkspacesAssociatedAssembly2008} also studied the five-bar, proposing a heuristic to move through the nearest output singularity to transition modes as necessary.

Much like a globe is a curved surface in 3D with shortest paths given by great circles, a five-bar configuration space is also a two degree-of-freedom surface in a higher dimensional \emph{ambient} space. In analogy with great circles, shortest paths between configurations 
are given by \emph{geodesics} on the surface. 
We conduct path planning and analyze the different modes of five-bar configuration spaces by modelling a configuration space as a weighted graph whose nodes are points in the configuration space and edges are weighted by distance in the ambient space. More precisely, since we aim to understand path planning while avoiding transmission problems, we must model the configuration space with input~\mbox{singularities~removed}. Through this data structure, we represent the input modes.

Using a weighted graph to model a workspace, configuration space, and, more generally, the geometry of a lower-dimensional manifold embedded in high-dimensional ambient space 
is a well-known strategy for both path planning and manifold learning (e.g., \cite{bohigas2012singularity,cyclotop2010,karaman2011sampling,latombe1999motion,tenenbaum2000global}). This is generally more computationally expensive than computing a single path between just a single pair of configurations (e.g., \cite{dalibard2009rrt,porta2012randomized}), but a robust graph model allows us to analyze more optimal paths and make repeated queries to investigate the geometry of output and input modes. The usefulness of any such graph model depends, however, on how closely the graph estimates the geometry of the space. 

In this paper, we present a method which uses numerical algebraic geometry \cite{BertiniBook,SW05} to compute an appropriate resolution for sampling~\cite{di2022computing,eklund2022numerical}, \emph{uniformly} (not randomly) sample a configuration space and construct a graph using that resolution~\cite{di2020sampling}, and avoid input singularities by computing distances to the input boundary~\cite{HauensteinReal,SeidenbergDistance}. In contrast to methods which randomly or iteratively build samples, our method uses algebraic computations to make a global choice for the sample and graph resolution. The distance objective used for path planning is close to optimal in the sense that it estimates geodesic distances in the configuration space. Singularity avoidance is quantified in terms of Euclidean distance, which is easier to relate to application considerations than, e.g., determinant based approaches. Since the only data required are polynomial equations defining the configuration space and input singularities, this method is applicable broadly.

\section{METHOD}

Let $C\subset\bR^n$ denote the $d$-dimensional 
configuration space which is a manifold\footnote{Recall that manifolds do not have self-intersections, cusps, or other geometric singularities.} in the \mbox{$n$-dimensional} ambient space
such that $C$ is defined by the system of $n-d$ polynomial equations
$\mbf{F}(\mbf{z}) = 0$ where $\mbf{z}\in\bR^n$.  
Suppose that the set of input singularities~$I$ in the configuration space
is defined by the polynomial equation
$g(\mbf{z})=0$ where $\mbf{z}\in C$.  For instance, consider the five-bar mechanism
as depicted in Fig.~\ref{fig:schematic}. The dimensions are given by~$a_x,a_y,b_x,b_y,l_1,l_2,l_3,l_4,p,q$. For canonical representation, $a_x=a_y=b_y=0$ can be assumed without loss of generality. The configuration space is defined by $\mbf{F}(\mbf{z})=0$:
\begin{align*} 
& x^2 + y^2 - 2l_1\left(xc_\phi+ys_\phi\right) + l_1^2 - p^2 - q^2 = 0,\nonumber\\
& l_2^2 (x^2 + y^2) + l_1^2 ((l_2 - p)^2 + q^2) + (b_x^2 + l_3^2 - l_4^2) (p^2 + q^2)\nonumber\\
&- 2 l_2 l_3 (p (x c_\psi + y s_\psi) - q (x s_\psi - y c_\psi))- 2 b_x l_2 (p x + q y)   \nonumber\\
&+ 2 l_1 l_3 ((l_2 p - p^2 - q^2) (c_\phi c_\psi + s_\phi s_\psi)  + 2 b_x l_3 (p^2 + q^2) c_\psi\nonumber\\
&+ 2 b_x l_1 ((l_2 p - p^2 - q^2) c_\phi + l_2 q s_\phi) \nonumber\\
&- 2 l_1 l_2 l_3 q (c_\phi s_\psi - s_\phi c_\psi) + 2 l_1 l_2 ((p - l_2) (x c_\phi + y s_\phi) \nonumber\\
& - q (x s_\phi - y c_\phi))= 0, \nonumber\\
& c_\phi^2 + s_\phi^2 - 1 = 0, \nonumber\\
& c_\psi^2 + s_\psi^2 - 1 = 0,
\end{align*}
where the variables are $\mbf{z} = (x,y,c_\phi,s_\phi,c_\psi,s_\psi)$ with $c_\phi,s_\phi$ and $c_\psi,s_\psi$ being the cosine and the sine of $\phi$ and $\psi$ in order, respectively. 
This replacement of angles with their corresponding cosine and sine 
along with the Pythagorean identity is a standard approach to yield polynomials
and it avoids redundancy related to the periodicity of angles.  
In particular, other than an increased number of variables and equations, 
this does not impact the results of path planning.  

Let $\mbf{F}_1$ and $\mbf{F}_2$ be the first two polynomial functions above. For the five-bar, the set of input singularities $I$ is defined by
\[
g(\mbf{z})  = \det\begin{pmatrix} \frac{\partial \mbf{F_1}}{\partial x}  & \frac{\partial \mbf{F_1}}{\partial y} \\[0.05in] 
                            \frac{\partial \mbf{F_2}}{\partial x}  & \frac{\partial \mbf{F_2}}{\partial y}
            \end{pmatrix} = 0
\hbox{~~~for~}\mbf{z}\in C.
\]
A first measure of the geometric complexity of $I$
is its degree in the ambient space.  Generically, for the five-bar,
$I$ is a union of two irreducible curves of degree $6$.
Note that these two curves arise from two four-bar coupler curves.

\subsection{Sampling and graph construction}

The graph construction will follow a straightforward radius graph approach.
For $\mbf{v}\in\bR^n$, let $\Vert \mbf{v} \Vert$ denote the standard Euclidean distance.
For \mbox{$r>0$}, the $r$-radius graph on a finite set $\hat{C}\subset\bR^n$, denoted $G_r(\hat{C})$, 
is the undirected graph with node set $\hat{C}$ and an edge $\mbf{e}=(\mbf{c_0},\mbf{c_1})$
that is weighted by distance $\Vert \mbf{c_0} - \mbf{c_1} \Vert$ for every pair of distinct points $\mbf{c_0}, \mbf{c_1}\in \hat{C}$ such that $\Vert \mbf{c_0} - \mbf{c_1} \Vert \leq r$. 

To model the configuration space $C$ with a radius graph, we require an appropriate finite sample\footnote{In fact, we can easily relax the requirement $\hat{C}\subset C$ to allow the set $\hat{C}$ to consist of points which are at most distance $\delta > 0$ from $C$ for small $\delta$. For the sake of clarity, we discuss the case where $\delta=0$.} 
$\hat{C}\subset C$ and also an appropriate radius $r > 0$. 
To be more precise regarding the sample, our method first determines an 
appropriate ``resolution'' $\epsilon>0$ and then computes 
an \emph{$\epsilon$-sample} of $C$, $\hat{C}$, such 
that every point $\mbf{c}\in C$ is within distance $\epsilon$ of a point in $\hat{C}$.

We can associate to $C$ two geometric feature sizes: the \emph{reach} of C \cite{FedererReach}, denoted
$\reach(C)$, and the \emph{weak feature size} of $C$ \cite{chazalwfs,Grove93Critical}, denoted $\wfs(C)$. 
These are numbers which, roughly speaking, quantify the size of the geometric features in $C$. 
They are related by $0 < \reach(C) \leq \wfs(C)$. 
See, e.g.,~\cite[Sec. 2]{di2022computing} for an expanded discussion. 
One can show (e.g.,~\cite[Thm 2.11]{di2022computing}) that if $2\epsilon < \wfs(C)$,
$\hat{C}$ is an $\epsilon$-sample of~$C$, 
and $r(\epsilon) = 4\epsilon\sqrt{\frac{2n}{n+1}}$, 
then the radius graph~$G_{r(\epsilon)}(\hat{C})$ 
has the same number of connected components as $C$. 
Moreover, if $2\epsilon < \reach(C)$, 
then for any edge $(\mbf{c_0},\mbf{c_1}) \in G_{r(\epsilon)}(\hat{C})$, 
there is a path in $C$ between $\mbf{c_0}$ and $\mbf{c_1}$
where the distance $\Vert \mbf{c_0} - \mbf{c_1} \Vert$ is a good estimate for the length of the path. 
The shortest path graph distance between two points in $G_{r(\epsilon)}(\hat{C})$ is subsequently a good estimate for the shortest path distance in $C$ between the points.

Numerical algebraic geometry methods which use the polynomial system $\mbf{F}$ to determine the reach and weak feature size of $C$ have recently been developed~\cite{di2022computing,eklund2022numerical}, as have methods which produce an $\epsilon$-sample of $C$ from $\mbf{F}$~\cite{di2020sampling,dufresne2019sampling}. One advantage of the latter method over, e.g., random sampling, is that we can use geometric subsampling methods to mitigate oversampling while maintaining an appropriate resolution.  The feature size computations are relatively expensive, so we use an early stopping criterion to estimate\footnote{Following the terminology of \cite{di2022computing}, we compute only the geometric 2-bottlenecks of $C$.} $\wfs(C)$. In summary, we construct a graph to model $C$ as follows:
\begin{enumerate}
    \item Compute an estimate $W$ for $\wfs(C)$ using the polynomial system $\mbf{F}$ as input.
    \item Compute an $\epsilon$-sample of $C$, $\hat{C}$, where $W=2\epsilon$, using the polynomial system $\mbf{F}$ as input.
    \item Construct the radius graph $G_{r(\epsilon)}(\hat{C})$.
\end{enumerate}

\subsection{Singularity avoidance}

An edge in this constructed graph corresponds with a path between two points
in the configuration space.  To avoid input singularities, one only needs, in theory, to remove edges which correspond to paths that cross the input boundary. Since we cannot check this directly, we instead
compute the distance in the ambient space between the set of input singularities
and the straight-line segment connecting an edge's two endpoints.  
This is justified by the heuristic that the straight-line
segment between the two points can be taken as a reasonable approximation 
of the geodesic in the configuration space connecting the two points.

Let $\mbf{c}_0,\mbf{c}_1\in C$ so that the straight-line segment connecting
them is $\mbf{c}(t) = (1-t) \mbf{c}_0 + t \mbf{c}_1$ for $t\in [0,1]$.
Hence, $\mbf{c}(0) = \mbf{c}_0$ and $\mbf{c}(1) = \mbf{c}_1$.
We aim to solve 
\begin{equation}\label{eq:MinDist}
\min \{\|\mbf{c}(t) - \mbf{w}\|^2 ~:~ \mbf{w} \in I, 0 \leq t \leq 1\}.
\end{equation}
Utilizing Fritz John first-order necessary conditions, one obtains 
the polynomial system $\mbf{J}(\mbf{w},t,\mbf{\lambda})$ equal to
$$\mbox{\small $
\left[\begin{array}{c}
\mbf{F}(\mbf{w}) \\
g(\mbf{w}) \\
\begin{array}{l}
\lambda_0 \nabla_{\mbf{w},t}\left(\|\mbf{c}(t) - \mbf{w}\|^2\right) + 
\sum_{i=1}^d \lambda_i \nabla_{\mbf{w},t}\left(\mbf{F}_i(\mbf{w})\right)\\
+\lambda_{d+1} \nabla_{\mbf{w},t}\left(g(\mbf{w})\right)+
\lambda_{d+2} \nabla_{\mbf{w},t}\left(t\right)+
\lambda_{d+3} \nabla_{\mbf{w},t}\left(1-t\right)
\end{array} \\
\lambda_{d+2} t \\
\lambda_{d+3}(1-t) 
\end{array}\right]$}$$
where $\mbf{\lambda}\in\mathbb{P}^{d+3}$
and $\nabla_{\mbf{w},t}(f(\mbf{w},t))$ is the gradient vector
of $f(\mbf{w},t)$ with respect to $\mbf{w}$ and $t$ for any function $f$.
 
Since we aim to solve $\mbf{J}=0$ for various choices of $\mbf{c}_0$ and $\mbf{c}_1$, 
we employ a parameter homotopy~\cite{CoeffParam} with
$\mbf{c}_0$ and $\mbf{c}_1$ as parameters.  
Thus, the first step is perform an {\em ab initio} solve of $\mbf{J}=0$
for generic values of the parameters.  
Note that the system $\mbf{J}$ has a natural
two-homogeneous structure with variables groups $(\mbf{w},t)$ and $\mbf{\lambda}$.
Hence, a multihomogeneous homotopy 
or a multihomogeneous regeneration~\cite{Regen,MhomRegen}
can be employed for this 
{\em ab initio} solve.
Then, a parameter homotopy simply deforms from the generic values of the parameter
to the given values of $\mbf{c}_0$ and $\mbf{c}_1$.  The number of solution
paths tracked in this parameter homotopy is precisely the generic number
of solutions to $\mbf{J}=0$.  
For the five-bar, the two-homogeneous B\'ezout count of $\mbf{J}$ is
$1152$ while the actual generic number of solutions to $\mbf{J}=0$
is $84$.  

The endpoints of the parameter homotopy correspond
with critical points of \eqref{eq:MinDist} for the 
given values of $\mbf{c}_0$ and $\mbf{c}_1$.
Hence, one can solve~\eqref{eq:MinDist} by sorting through the critical
points to determine real critical points on the line segment,
i.e., for $t\in[0,1]$, and selecting the corresponding minimum distance.
Moreover, by considering the last two polynomials in $\mbf{J}$,
which correspond to the complimentary slackness condition,
one can actually break the computation into three separate pieces:
at $t=0$, at $t=1$, and along the interior of the line segment, i.e., for $0<t<1$.  
Corresponding parameter homotopies would track $24$, $24$, and $36$ paths, respectively,
with $24+24+36=84$.  Since many edges in the graph can have the same node, 
one advantage of breaking this computation into these three separate pieces
is to avoid recomputing the same information for repeated nodes.

Performing this computation for every edge $\mbf{e}$ in the constructed graph results in a minimum distance $D(\mbf{e})$ along that edge to the input singularity and a distance $D(\mbf{c})$ for every configuration $\mbf{c}$ in the node set $\hat{C}$. To use the graph for singularity avoidance, i.e., to model the space $C\setminus I$, we then choose a distance threshold $T$ and remove all edges from the graph with $D(\mbf{e}) < T$. Any remaining nodes with no edges are also removed. 
In principle, one would like to compute a threshold $T$ such that the reduced graph has as many connected components as input modes. This would require a distinct and currently infeasible feature size computation for~$I$ as a subspace of the manifold $C$. 
At the present, we instead choose $T = r(\epsilon)$ based on simple geometric heuristics. 
Note, however, that we can also choose~$T$ based on engineering considerations. 
Singularity-avoiding path planning queries can be subsequently performed by shortest path computations in this graph, e.g., with $A^*$ path planning~\cite{hart1968formal} using the Euclidean distance 
in ambient space between two configurations as a heuristic. 

\subsection{Computational considerations}
The most computationally expensive steps of this method are 
computing the $\epsilon$-sample of $C$, $\hat{C}$, 
and computing the input boundary distance $D(\mbf{e})$ for each edge $\mbf{e}$ in the graph. 
Both of these steps can, however, take advantage of CPU parallelization, 
the former by utilizing parallelization in numerical algebraic geometry computations, 
the latter by computing distances for each edge in parallel. 
Also note that, after computing minimum distances from the graph nodes to~$I$, 
one may use that information to forego computing $D(\mbf{e})$ for any edge $\mbf{e}$
with endpoints sufficiently far away from $I$ relative to the threshold $T$. 

\section{RESULTS}\label{Sec:Examples}

The methodologies described above were applied to two examples of nonsymmetric five-bar linkages.  Their parameters are listed in~Table~\ref{tab:Parameters}
with
Table~\ref{tab:graph_info} summarizing 
the size of the graphs computed.
Both examples involve traversing across non-neighboring output modes while maintaining the same input mode.
In other words, output singularities are crossed and input singularities are not.



\begin{table}
    \vspace{1.5mm}
    \centering
    \begin{tabular}{r|r|r|r|r|r}
         Case & \multicolumn{1}{|c|}{$a_x$} & 
         \multicolumn{1}{|c|}{~$b_x$}& 
         \multicolumn{1}{|c|}{$p$}
         & 
         \multicolumn{1}{|c|}{$l_1$}& 
         \multicolumn{1}{|c}{$l_2$}\\
         \hline
         \multicolumn{1}{c|}{1} & 0.259 & 0.060 & 0.049 & 0.465 & 0.349 \\
         \multicolumn{1}{c|}{2} & 0.066 & $-$0.642 & 0.298 & 0.775 & 0.832  \\
         \\
         Case & 
         \multicolumn{1}{|c|}{$a_y$}& 
         \multicolumn{1}{|c|}{~$b_y$} & 
         \multicolumn{1}{|c|}{$q$} & \multicolumn{1}{|c|}{$l_3$} &  
         \multicolumn{1}{|c}{$l_4$} \\
         \hline
         \multicolumn{1}{c|}{1} & 0.586 & 0.590 & 0.328 & 0.249 & 0.411\\
         \multicolumn{1}{c|}{2} & 0.815 & 0.845 & 1.304 & 0.291 & 0.522
    \end{tabular}
    \caption{Five-bar dimensions for the two cases considered}
    \label{tab:Parameters}
\end{table}

\begin{table}
    \vspace{1mm}
    \centering
    \begin{tabular}{c|c|c}
         Case & \# pts in graph & \# edges in graph \\
         \hline
         1 & \multicolumn{1}{|r|}{82,581~~~~} & $2.3\times 10^6$  \\
         2 & \multicolumn{1}{|r|}{121,667~~~~} & $3.2\times 10^6$   \\
    \end{tabular}
    \caption{Summary of the radius graphs for the two cases considered}
    \label{tab:graph_info}
\end{table}

\begin{figure}[htbp]
    \centering
	\includegraphics[width=0.35\textwidth]{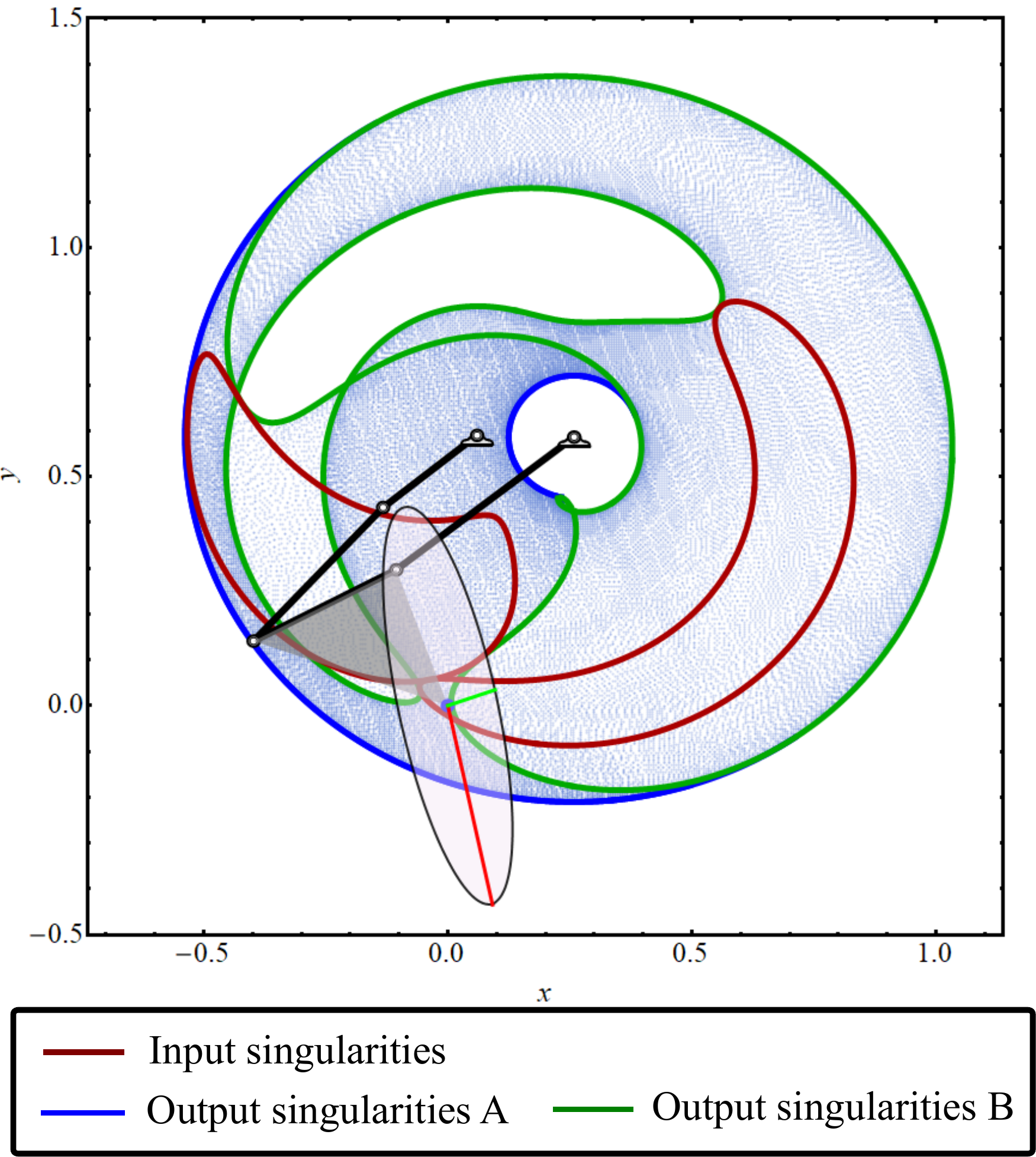}
	\caption{The workspace for Case Study~1.  At the $x$-$y$ point shown exists another configuration with a velocity ellipse perpendicular to the one shown.}
	\label{fig:workspace1}
\end{figure}

\begin{figure}[htbp]
    \centering
	\includegraphics[width=0.425\textwidth]{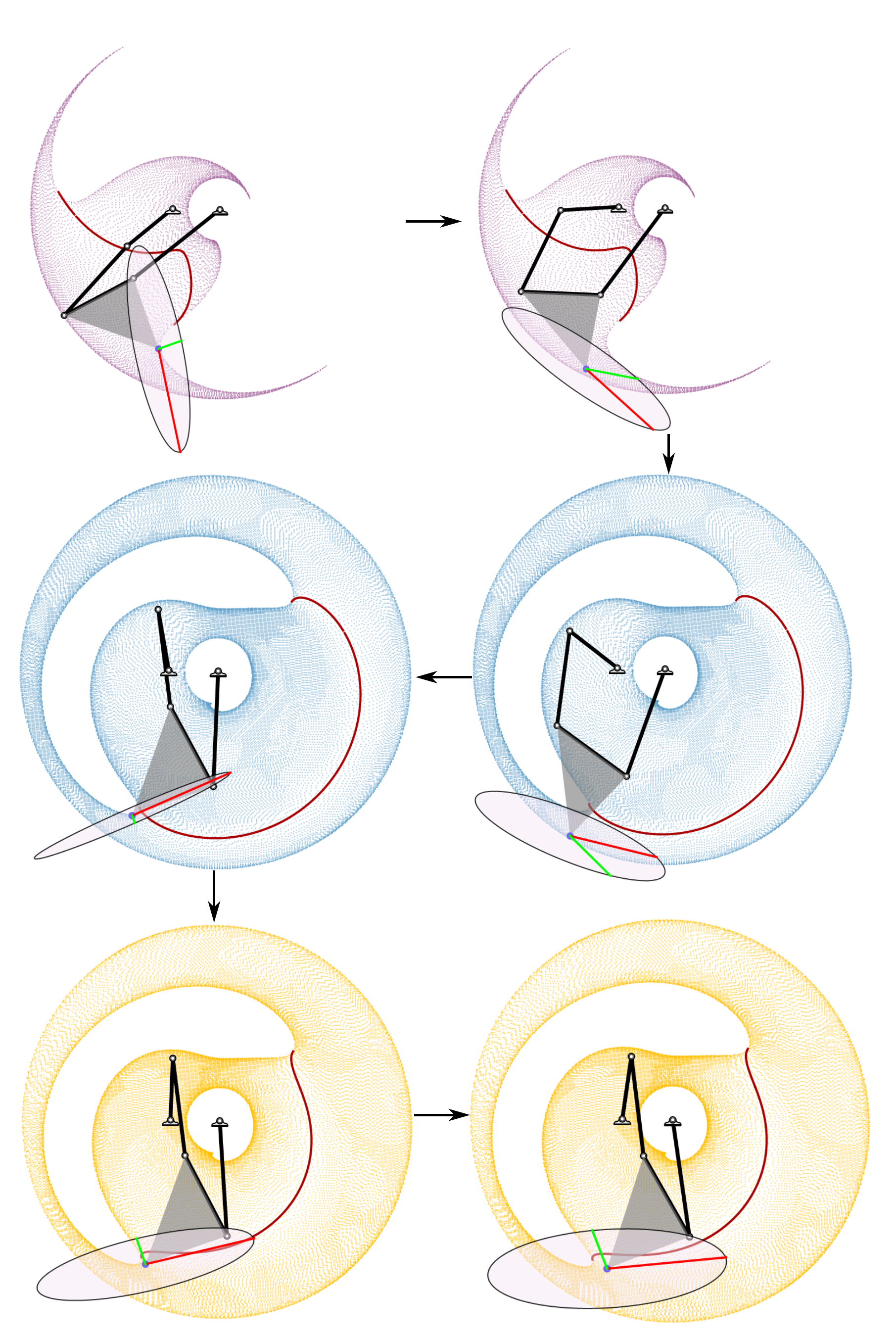}
	\caption{Case Study~1: A feasible path marked with six snapshots to switch between the two perpendicular velocity ellipses. The workspace and the input singularities curves shown correspond to the output mode to which each configuration~belongs.}
	\label{fig:case1}
\end{figure}

\begin{figure}[htbp]
    \vspace{1.5mm}
    \centering
	\includegraphics[width=0.35\textwidth]{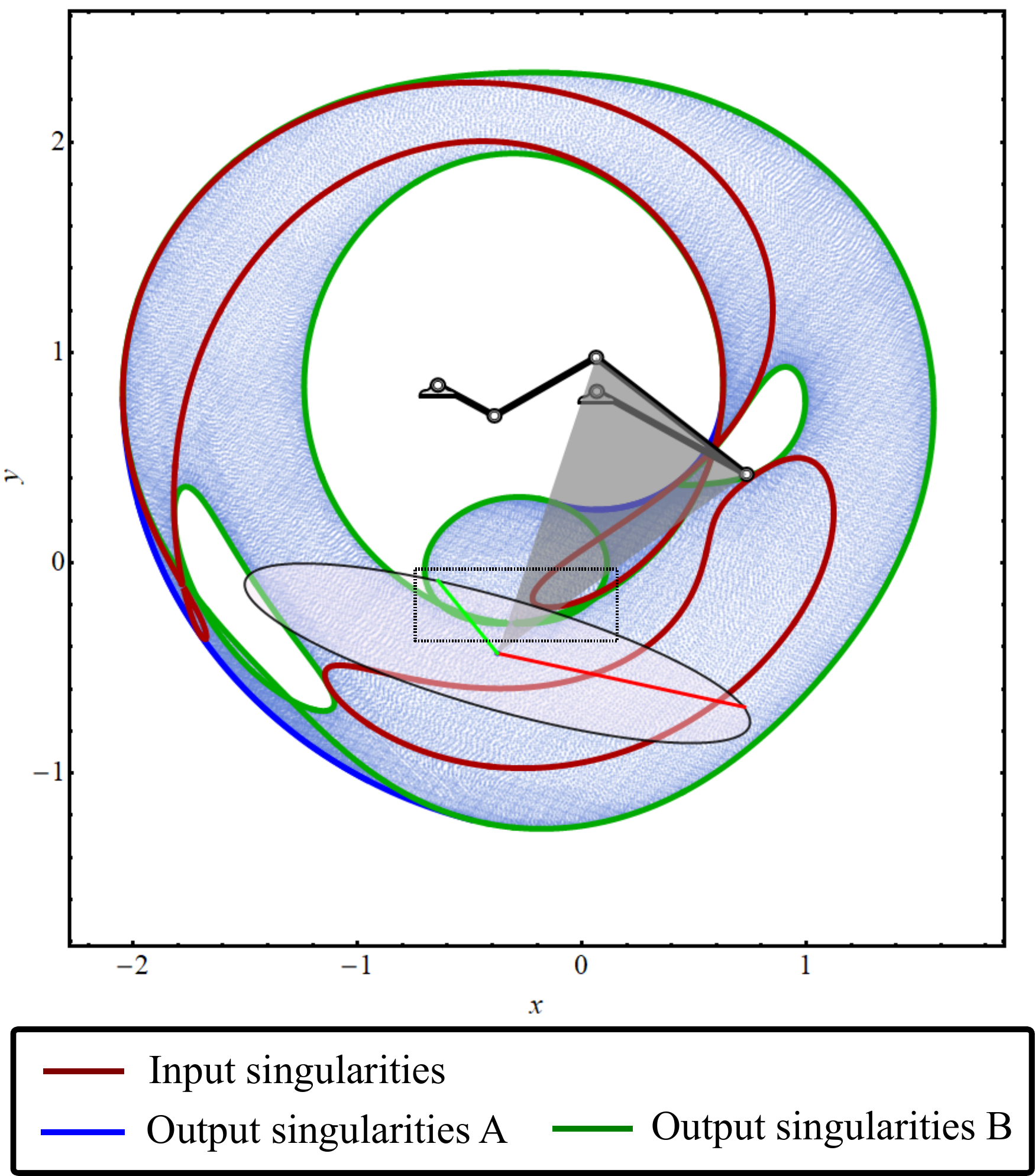}
	\caption{The workspace for Case Study~2.  The portion of the output bound which functions as a ceiling or floor is indicated with a box.}
	\label{fig:workspace2}
\end{figure}

\begin{figure}[htbp]
    \centering
	\includegraphics[width=0.425\textwidth]{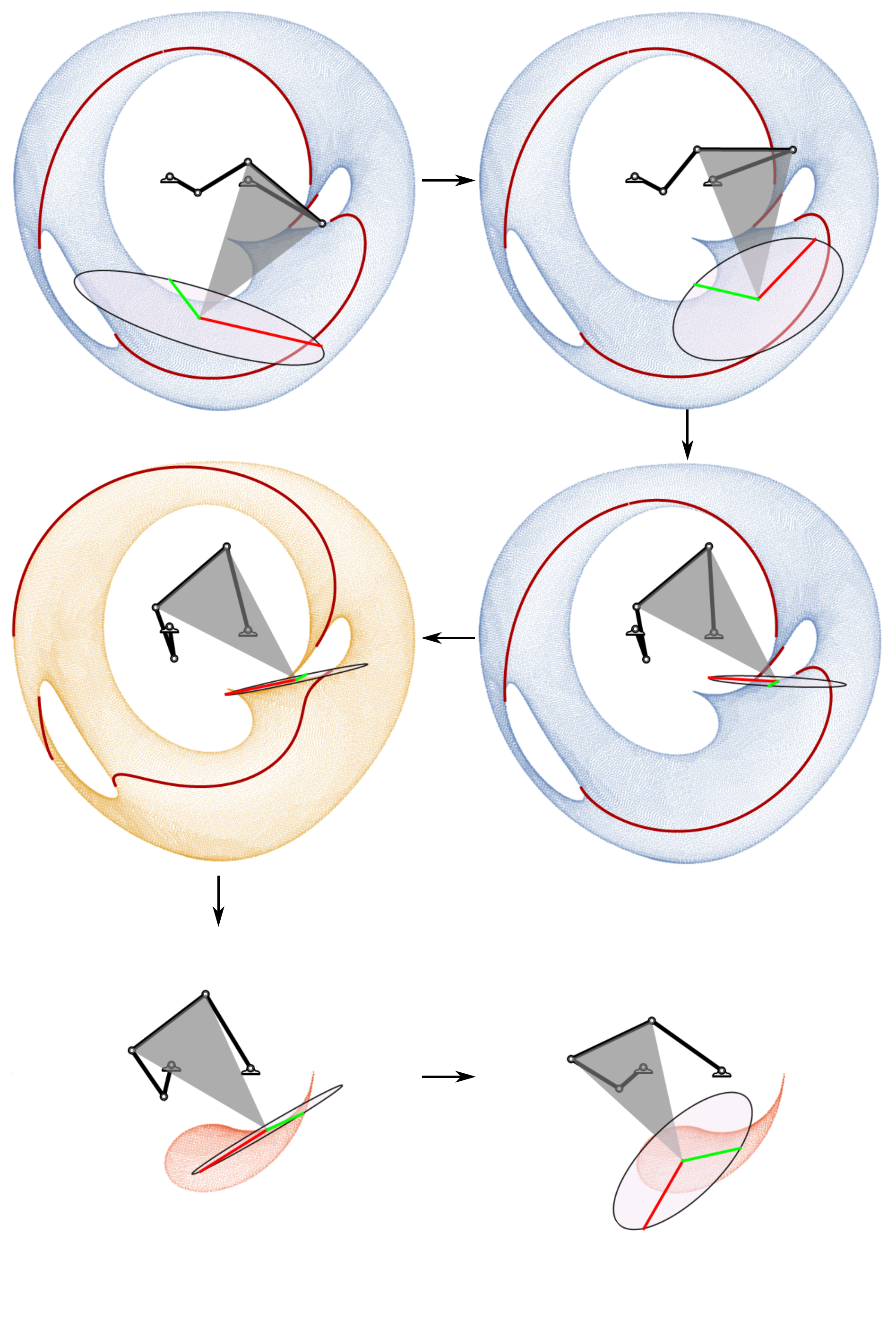}
	\vspace{-8mm}
	\caption{Case Study~2: A feasible path marked with six snapshots for the ceiling-to-floor example. The workspace and the input singularities curves shown correspond to the output mode to which each configuration~belongs.}
	\label{fig:case2}
	\vspace{-4mm}
\end{figure}

\subsection*{Case study 1: Perpendicular ellipses at a shared work\-space~point}

The first example five-bar possesses a point in its workspace where two of the four IKP solutions exhibit velocity ellipses with a 4:1 aspect ratio that are perpendicular to each other.
Its dimensions were found using the method reported in \cite{plecnikEllipseSynthesisFivebar2022}.
The direction of the short axis enjoys 4x larger force transmission while the direction of the long axis can achieve 4x greater speeds.    
The configurations corresponding to these perpendicular ellipses are housed in different output modes.
The path planning challenge is to approximate the geodesic from one to the other, which in this case involves transitioning through at least two~output~singularities.

At the computed singularity avoidance threshold, the graph model of the configuration space estimated 6 input modes, 5 output modes, and 7 input/output modes (maximal regions bounded by both input and output singularities). 
Due to the numerical threshold, our counts generally do not match the true number of regions, as they partition the configuration space with a thicker kerf, so to say.
In some respects, this is more practical as input singularities should be avoided by a margin of safety.
To give some idea of the space's extrinsic curvature, the distance in ambient space between the two configurations of interest is 2.11, the shortest path in the configuration space estimated by our method without input singularity avoidance is 2.66, and the shortest point with input singularity avoidance is 2.74.


\subsection*{Case study 2: A ceiling and a floor}

The second example five-bar possesses an output mode where there exists an arc that serves as a local upper bound for the $y$-coordinate of the end-effector, referred to as a \emph{ceiling}.
The same five-bar also possesses another output mode where there exists another arc, nearly on top of the first arc, but now functioning as a local lower bound for the $y$-coordinate of the end-effector, referred to as a \emph{floor}.
The path planning challenge is to transition from the output mode with a ceiling to the output mode with a floor.
This could be useful because, at an output bound, the mechanism is able to support large loads in the bound's normal direction without any actuator effort.
However, as it is a bound, regular motions can only take place on one side, locally.
However, by transitioning between a ceiling and a floor, the side of regular motion becomes selectable.

Our method estimated 5 input modes, 3 output modes, and 12 input/output modes. The ambient distance between the configuration in Fig.~\ref{fig:case2} was 2.61 and the path distance both with and without singularity avoidance was 4.82. The two 
paths distances are the same since the computed path does not come particularly close to an input singularity.


\section{CONCLUSIONS}

This paper presented a motion planning technique for transitioning between points in the configuration space of a five-bar manipulator.
This is a challenging task since the input and output singularities of parallel manipulators appear in a complicated manner.  
To overcome this, we constructed a radius graph to represent the workspace with edges weighted by nearness to input singularities.
The graph is partitioned into input modes and a path planner is employed to approximate geodesics between start and final points that avoid input singularities.
The resulting algorithm is able to traverse through multiple output modes in 
a smooth manner as demonstrated with two cases.

\addtolength{\textheight}{-6.5cm}   








\input{main.bbl}

\end{document}